\documentclass[11pt]{article}

\usepackage[final]{acl}

\usepackage{times}
\usepackage{latexsym}

\usepackage[T1]{fontenc}

\usepackage[utf8]{inputenc}

\usepackage{microtype}

\usepackage{inconsolata}

\usepackage{graphicx}
\usepackage{multirow}
\usepackage{amssymb}

\usepackage{booktabs}
\usepackage{multirow}
\usepackage{xcolor}
\usepackage{pifont}
\usepackage{amsmath}
\usepackage{colortbl}
\usepackage{adjustbox}
\usepackage{array}

\usepackage{makecell}
\usepackage{tabularx}
\usepackage{caption}
\usepackage{subcaption}

\usepackage[most]{tcolorbox}
\usepackage{fvextra} 

\tcbset{
  qapromptbox/.style={
    enhanced,
    breakable,
    colback=black!2,
    colframe=black!45,
    boxrule=0.4pt,
    arc=2pt,
    left=6pt,right=6pt,top=4pt,bottom=4pt,
    title={QA Generation Prompt},
    fonttitle=\bfseries,
  }
}

\definecolor{mygreen}{RGB}{0,120,0}
\definecolor{myred}{RGB}{170,0,0}
\definecolor{mutedblue}{RGB}{45,90,160}
\definecolor{mutedred}{RGB}{170,70,70}

\newcommand{\cmark}{\textcolor{mygreen}{\ding{51}}} 
\newcommand{\xmark}{\textcolor{myred}{\ding{55}}}   

\title{Document-Level Numerical Reasoning across\\Single and Multiple Tables in Financial Reports}

\author{
\textbf{Yi-Cheng Wang}\thanks{Equal contribution.}$^{1}$ \quad
\textbf{Wei-An Wang}\footnotemark[1]$^{1,2}$ \quad
\textbf{Chu-Song Chen}$^{1,2}$ \\
$^1$Department of Computer Science and Information Engineering, National Taiwan University \\ 
$^2$FinTech Center, National Taiwan University \\
\texttt{\{d13922033, r12922114, chusong\}@csie.ntu.edu.tw}
}

\begin{document}
\maketitle

\begin{abstract}
Despite the strong language understanding abilities of large language models (LLMs), they still struggle with reliable question answering (QA) over long, structured documents, particularly for numerical reasoning. Financial annual reports exemplify this difficulty: financial statement analysis often hinges on accurate arithmetic, and analysts derive key indicators by integrating evidence scattered across multiple tables and narrative text. However, existing benchmarks focus largely on single-table settings, leaving cross-table document-level numerical reasoning underexplored. To address this gap, we introduce \textbf{FinLongDocQA}, a dataset for both single-table and \textit{cross-table} financial numerical reasoning in long-context reports. Evaluating both closed-source and open-source LLMs on FinLongDocQA reveals two bottlenecks: (1) annual reports often exceed 129k tokens, exacerbating the \emph{context rot} problem for locating relevant tables; and (2) even when relevant evidence is located, LLMs remain prone to errors in multi-step numerical reasoning. We propose \textbf{FinLongDocAgent}, a Multi-Agent Multi-Round Retrieval-Augmented Generation (RAG) approach that iteratively retrieves evidence, performs intermediate calculations, and verifies results across rounds. Experiments highlight the importance of iterative retrieval and verification for reliable numerical QA in long financial documents. Our dataset and code are available on \href{https://github.com/AI-Application-and-Integration-Lab/FinLongDocQA}{\textit{https://github.com/AI-Application-and-Integration-Lab/FinLongDocQA}}.

\end{abstract}

\section{Introduction}

Despite recent progress in LLMs~\cite{openai2025gpt52systemcard, google_gemini3, yang2025qwen3, liu2025deepseek}, building systems that reliably perform human-level reasoning remains challenging, particularly when decisions require integrating information from heterogeneous sources. Finance is a representative domain of this challenge, as many analytical tasks depend on combining numerical and textual evidence from complex documents. 

Financial numerical reasoning~\cite{chen-etal-2021-finqa, zhu-etal-2021-tat} refers to the ability to interpret numerical values, perform arithmetic operations, and derive meaningful ratios and trends. Many financial analysis tasks can be decomposed into such operations, including trend analysis, margin computation, and liquidity ratio estimation. While these tasks may appear straightforward, performing numerical reasoning over long-form financial reports is challenging even for \textbf{\textit{human analysts}}, as it requires identifying relevant tables across different sections and linking numerical values with associated textual descriptions. This process often involves iterative cross-referencing, making it both time-consuming and cognitively demanding.

\begin{figure*}
    \centering
    \includegraphics[width=1\linewidth]{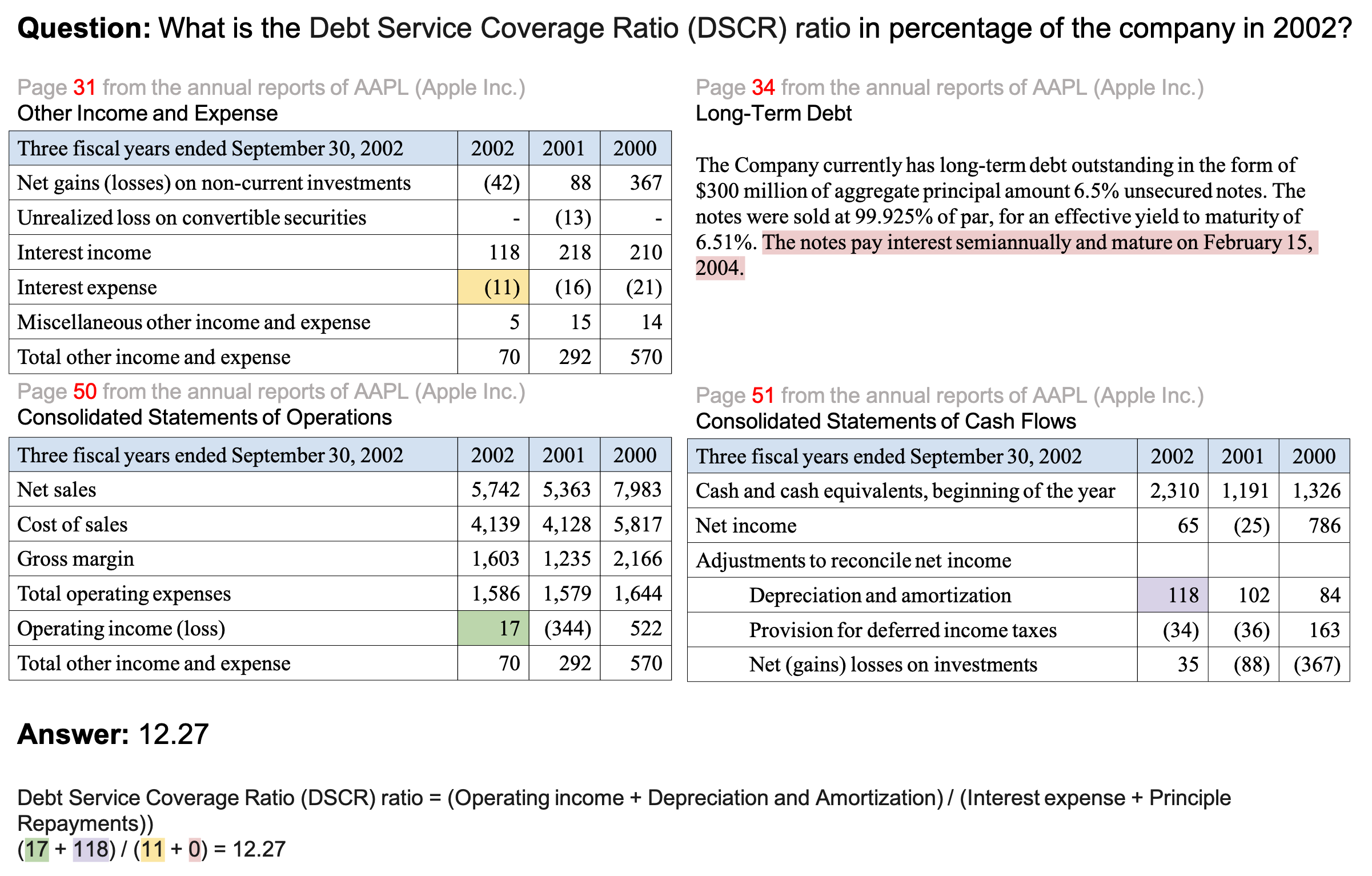}
    \caption{An example QA instance from FinLongDocQA. The figure shows only the relevant tables and text for presentation; in practice, the model must retrieve them from the full annual report before computing the answer.}
    \vspace{-3pt}
    \label{fig:finlongdocqa_example}
\end{figure*}

Figure~\ref{fig:finlongdocqa_example} shows a example: computing the Debt Service Coverage Ratio (DSCR) requires combining operating income from the income statement (Page~50) with depreciation and amortization from the cash flow statement (Page~51), while also retrieving interest expense from an income/expense table (Page~31) and principal repayment details from a long-term debt note in the text span (Page~34). As a result, the core difficulty is not only performing arithmetic, but also locating the correct evidence across sections and heterogeneous formats (e.g., tables and text).

However, existing benchmarks largely focus on cross-text or single-table settings. As summarized in Table~\ref{tab:dataset_comparison}, early financial datasets such as Financial PhraseBank~\cite{10.1002/asi.23062} and FiQA-SA~\cite{10.1145/3184558.3192301} target text-only semantic analysis, while numerical reasoning benchmarks such as FinQA~\cite{chen-etal-2021-finqa} and TAT-QA~\cite{zhu-etal-2021-tat} associate each question with a single table and a passage. More recent RAG-oriented benchmarks (e.g., FinDER~\cite{10.1145/3768292.3770361} and FinTMMBench~\cite{10.1145/3746027.3755723}) broaden task coverage and incorporate additional modalities, but they do not specifically target document-level numerical reasoning that requires retrieving and aggregating evidence across multiple tables within a long report. This leaves a gap for multi-table, multi-text workflows that are common in practical financial analysis.



Motivated by these challenges and the limitations of existing datasets, we introduce \textbf{FinLongDocQA}, a document-level dataset designed for both single-table and cross-table financial numerical reasoning in long-context settings. FinLongDocQA is constructed from S\&P 500 companies' annual reports from 2022$\sim$2024, and consists of 1,456 reports and 7,527 QA pairs, with an average input length of approximately 129k tokens.

Due to the structured nature of financial reports, meaningful numerical questions often rely on tables located within nearby pages, where related information is organized into coherent sections. While long-distance dependencies do exist, they are less frequent in practice. During dataset construction, we observed that cases requiring excessively remote evidence often led to ambiguous questions and were therefore excluded to ensure quality and answerability. Notably, our dataset still includes 1,280 questions whose supporting evidence spans over 40 pages on average, with a maximum span of 607 pages. This scale and level of long-range evidence aggregation exceeds prior benchmarks such as DocFinQA~\cite{reddy-etal-2024-docfinqa}, which focus on single-table evidence and therefore do not involve cross-page, multi-table reasoning.

We use FinLongDocQA to evaluate both closed-source (e.g., GPT~\cite{achiam2023gpt} and Gemini~\cite{google_gemini3}) and open-source (e.g., Qwen3~\cite{yang2025qwen3} and DeepSeek-v3.2~\cite{liu2025deepseek}) LLMs. Our findings indicate two key challenges. First, the extreme length and structure of annual reports make it difficult for models to reliably locate relevant evidence, especially when supporting tables are distributed across distant pages, reflecting a \emph{context rot}~\cite{hong2025context, liu-etal-2024-lost} problem. Second, even when relevant evidence pages are retrieved, models remain error-prone in multi-step numerical reasoning~\cite{10.1145/3677052.3698682}, including extracting incorrect table values, misapplying formulas, and mixing units. These observations suggest that solving FinLongDocQA requires both robust evidence discovery and reliable computation over retrieved content.

To establish a strong baseline for document-level cross-table numerical QA, we introduce \textbf{FinLongDocAgent}, a Multi-Agent Multi-Round RAG method. FinLongDocAgent decomposes the task into iteratively refines retrieval and reasoning: agents generate domain-aware queries to retrieve candidate tables and text spans, perform intermediate calculations, and verify results across multiple rounds. This iterative design aims to improve evidence coverage in long documents and enhance numerical reliability in cross-table reasoning.

Our contributions are summarized as follows.

\noindent $\bullet$~We create \textbf{FinLongDocQA}, the first benchmark (to our knowledge) targeting document-level financial numerical QA that requires reasoning across \textit{multiple tables} within long-form annual reports.

\noindent $\bullet$~We provide page-grounded evidence annotations and executable programs for every instance, enabling fine-grained evaluation.

\noindent $\bullet$~We introduce \textbf{FinLongDocAgent}, an agent-based RAG method, and provide detailed analysis identifying key challenges of FinLongDocQA.

\begin{table}[t]
    \centering
    \small
    \setlength{\tabcolsep}{3pt}
    \resizebox{\linewidth}{!}{
\begin{tabular}{llccccc} 
\toprule
\multirow{2}{*}{\textbf{Dataset}} & \multirow{2}{*}{\textbf{Task}} & \multicolumn{3}{c}{\textbf{Input Modality}} & \multirow{2}{*}{\begin{tabular}[c]{@{}c@{}}\textbf{Doc}\\\textbf{RAG}\end{tabular}} & \multirow{2}{*}{\begin{tabular}[c]{@{}c@{}}\textbf{Cross}\\\textbf{Table}\end{tabular}} \\ 
\cmidrule(l){3-5}
 &  & \textbf{Tab.} & \textbf{Text} & \textbf{Img.} &  &  \\ 
\midrule
FiQA-SA (WWW'18) & Semantic analysis & \xmark & \cmark & \xmark & \xmark & \xmark \\
FinQA (EMNLP'21) & Numerical reasoning & \cmark & \cmark & \xmark & \xmark & \xmark \\
TAT-QA (ACL'21) & Numerical reasoning & \cmark & \cmark & \xmark & \xmark & \xmark \\
TAT-DQA (MM'22) & Numerical reasoning & \xmark & \cmark & \cmark & \xmark & \xmark \\
MultiHiertt (ACL'22) & Numerical reasoning & \cmark & \cmark & \xmark & \xmark & \cmark \\
DocFinQA (ACL'24) & Numerical reasoning & \cmark & \cmark & \xmark & \cmark & \xmark \\
DocMath-Eval (ACL'24) & Numerical reasoning & \cmark & \cmark & \xmark & \cmark & \xmark \\
AlphaFin (LREC'24) & Stock prediction & \xmark & \cmark & \xmark & \cmark & \xmark \\
CodeFinQA (ACL'24) & Code generation & \cmark & \cmark & \xmark & \xmark & \xmark \\
FinanceMath (ACL'24) & Math reasoning & \cmark & \cmark & \xmark & \cmark & \xmark \\
FinDER (ICAIF'25) & Retrieval & \cmark & \cmark & \xmark & \cmark & \xmark \\
FinTMMBench (MM'25) & Mix (Include 10 tasks) & \cmark & \cmark & \cmark & \cmark & \xmark \\
FinMMR (ICCV'25) & Numerical reasoning & \cmark & \cmark & \cmark & \xmark & \xmark \\
XFinBench (ACL'25) & Complex reasoning & \cmark & \cmark & \cmark & \xmark & \xmark \\
FCMR (ACL'25) & Cross-modal reasoning & \cmark & \cmark & \cmark & \cmark & \xmark \\
FinanceReasoning (ACL'25) & Numerical reasoning & \cmark & \cmark & \xmark & \xmark & \xmark \\ 
\midrule
\textbf{FinLongDocQA (ours)} & Numerical reasoning & \cmark & \cmark & \xmark & \cmark & \cmark \\
\bottomrule
\end{tabular}
}
\caption{Comparison of financial QA datasets. \textbf{Doc RAG} denotes settings where QA is grounded in a full document and retrieval is required to find evidence.}
\label{tab:dataset_comparison}
\vspace{-10pt}
\end{table}

\section{Related Work}
\noindent\textbf{Financial QA Benchmarks.}
Early benchmarks such as Financial PhraseBank~\cite{10.1002/asi.23062} and FiQA-SA~\cite{10.1145/3184558.3192301} mainly target sentiment analysis and opinion mining in financial text. FinQA~\cite{chen-etal-2021-finqa} and TAT-QA~\cite{zhu-etal-2021-tat} shift to numerical QA over financial reports in a given-evidence setting, where models perform arithmetic over provided table and text snippets. MultiHiertt~\cite{zhao-etal-2022-multihiertt} extends this setting to multiple tables, increasing the need for multi-step reasoning. 

Recent benchmarks broaden the scope in two directions. First, several datasets incorporate visual and layout information from documents and charts, including TAT-DQA~\cite{10.1145/3503161.3548422}, FinMMR~\cite{Tang_2025_ICCV}, FCMR~\cite{kim-etal-2025-fcmr}, and FinTMMBench~\cite{10.1145/3746027.3755723}. Second, benchmarks such as AlphaFin~\cite{li-etal-2024-alphafin} and XFinBench~\cite{zhang-etal-2025-xfinbench} evaluate a wider range of financial analysis skills beyond report QA. To better assess calculation correctness, FinanceMath~\cite{zhao-etal-2024-knowledgefmath}, BizBench~\cite{krumdick-etal-2024-bizbench}, and FinanceReasoning~\cite{tang-etal-2025-financereasoning} emphasize formula-based reasoning and program-style solutions (e.g., Program-of-Thought~\cite{chen2023program}). 

Work on long-context financial QA, such as DocFinQA~\cite{reddy-etal-2024-docfinqa}, DocMath-Eval,  and FinDER~\cite{10.1145/3768292.3770361}, highlights evidence localization challenges in lengthy documents. However, these settings do not require combining evidence across multiple tables dispersed throughout an entire report. In practice, financial statement analysis often requires aggregating values across several tables and accompanying text, motivating our focus on document-level cross-table numerical reasoning.

\begin{figure}[t]
    \centering
    \includegraphics[width=1\linewidth]{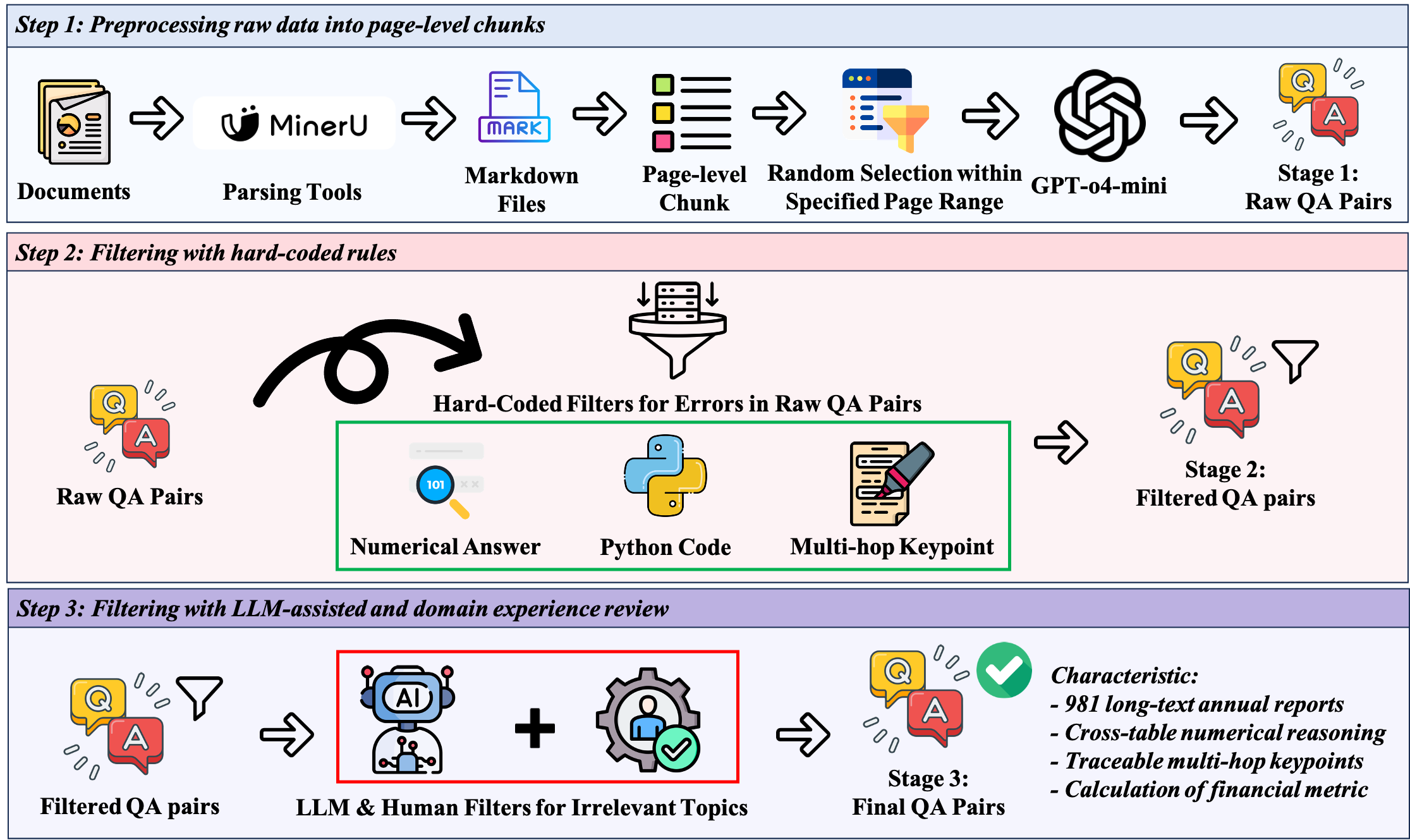}
    \caption{Illustration of the construction steps for our FinLongDocQA dataset.}
    \label{fig:dataset_pipeline}
\end{figure}

\noindent\textbf{Cross-Table Reasoning.}
Multiple benchmarks in the general domain evaluate cross-table reasoning. MMQA~\cite{wu2025mmqa} targets multi-hop QA over relationally linked tables with explicit primary/foreign keys, while TableBench~\cite{10.1609/aaai.v39i24.34739} and LongTableBench~\cite{li-etal-2025-longtablebench} assess table understanding and long-context reasoning across diverse non-financial domains. Financial reports pose a different challenge: table relations are typically implicit and must be inferred from reporting structure and narrative references rather than database-style keys. In contrast to these general-domain benchmarks, our proposed FinLongDocQA requires integrating evidence from multiple tables dispersed throughout a long annual report.

\noindent\textbf{Retrieval-Augmented Generation (RAG).}
Recent advances in LLMs have focused on long-context reasoning and explicit ``thinking'' capabilities. Closed-source models such as GPT-5.2~\cite{openai2025gpt52systemcard} and Gemini-3~\cite{google_gemini3}, as well as open-source models like Qwen3~\cite{yang2025qwen3} and DeepSeek-V3.2~\cite{liu2025deepseek}, generate reasoning traces to handle complex tasks. In the financial domain, Fin-o1~\cite{qian2025fino1} further enhances reasoning by training on specialized financial corpora with reinforcement learning (RL). However, LLMs remain prone to \textit{context rot}~\cite{hong2025context, liu-etal-2024-lost} failures when locating relevant evidence in long-form documents.

RAG mitigates this issue by retrieving compact sets of relevant evidence instead of processing the entire context. In finance, FinQAPT~\cite{10.1145/3677052.3698682} optimizes context extraction for the FinQA dataset but reports performance degradation when scaled to full-length reports, and its system is not publicly released. To better capture complex dependencies in general-domain corpora, graph-based RAG approaches such as GraphRAG~\cite{edge2024local}, LightRAG~\cite{guo-etal-2025-lightrag}, and HippoRAG~2~\cite{gutierrez2025from} build graph indices over entities and relations, enabling global reasoning across documents.

More recently, research has moved toward agentic RAG, where models autonomously plan and execute search trajectories. ReAct~\cite{yao2022react} pioneered the combination of reasoning traces with task-specific actions. Subsequent work, such as Search-R1~\cite{jin2025searchr} and WebDancer~\cite{wu2025webdancer} uses RL to optimize multi-turn search interactions for information seeking.  While these agents are effective for retrieval and general search, they are not designed for the iterative numerical verification required in cross-table analysis. 


\begin{table}[t]
\centering
\small
\setlength{\tabcolsep}{6pt}
\renewcommand{\arraystretch}{1.08}

\begin{adjustbox}{max width=\columnwidth}
\begin{tabular}{p{0.78\linewidth}r}
\toprule
\textbf{Stage} & \textbf{\# QA pairs} \\
\midrule
Raw data generated by LLM & 13{,}375 \\
\midrule
\textit{Rule-based filtering stage} & \\
\quad Removal of single-table solvable cases & 4{,}120 \\
\quad Removal of calculation errors & 657 \\
\quad Removal of incoherent answers & 268 \\
\midrule
Manual review filtering stage & 803 \\
\midrule
\textbf{Final dataset} & \textbf{7{,}527} \\
\bottomrule
\end{tabular}
\end{adjustbox}
\caption{Dataset construction and filtering statistics.}
\label{tab:dataset_filtering}
\end{table}

\begin{table}[t]
\centering
\small
\setlength{\tabcolsep}{6pt}
\renewcommand{\arraystretch}{1.08}

\begin{adjustbox}{max width=\columnwidth}
\begin{tabular}{p{0.72\linewidth}r}
\toprule
\textbf{Statistic} & \textbf{Value} \\
\midrule
\multicolumn{2}{l}{\textit{Documents}} \\
\quad \#Reports & 1{,}456 \\
\quad \#Companies & 489 \\
\quad Tables & 150{,}251 \\
\quad Tokens (mean/median) & 129{,}589 / 115{,}182 \\
\midrule
\multicolumn{2}{l}{\textit{Question-answer pairs}} \\
\quad Total QA pairs & 7{,}527 \\
\bottomrule
\end{tabular}
\end{adjustbox}
\caption{FinLongDocQA dataset overview.}
\label{tab:dataset_overview}
\end{table}

\begin{table}[t]
\centering
\small
\setlength{\tabcolsep}{6pt}
\renewcommand{\arraystretch}{1.08}
\begin{adjustbox}{max width=\columnwidth}
\begin{tabular}{p{0.72\linewidth}r}
\toprule
\textbf{Statistic} & \textbf{Value} \\
\midrule
\multicolumn{2}{l}{\textit{Question types}} \\
\quad Mixed / Table / Text & 5{,}951 / 1{,}319 / 257 \\
\multicolumn{2}{l}{\textit{Question length}} \\
\quad Tokens (mean/median) & 15.31 / 15 \\
\quad Characters (mean/median) & 92.41 / 90 \\
\midrule
\multicolumn{2}{l}{\textit{Evidence}} \\
\quad Evidence pages (mean/median/min/max) & 2.02 / 2 / 2 / 4 \\
\multicolumn{2}{l}{\textit{Annotations \& code}} \\
\quad Code lines (mean/median/max) & 3.02 / 3 / 12 \\
\bottomrule
\end{tabular}
\end{adjustbox}
\caption{Question, evidence, and annotation statistics in FinLongDocQA.}
\label{tab:dataset_qa_stats}
\end{table}

\begin{figure}[t]
  \centering
  \includegraphics[width=\columnwidth]{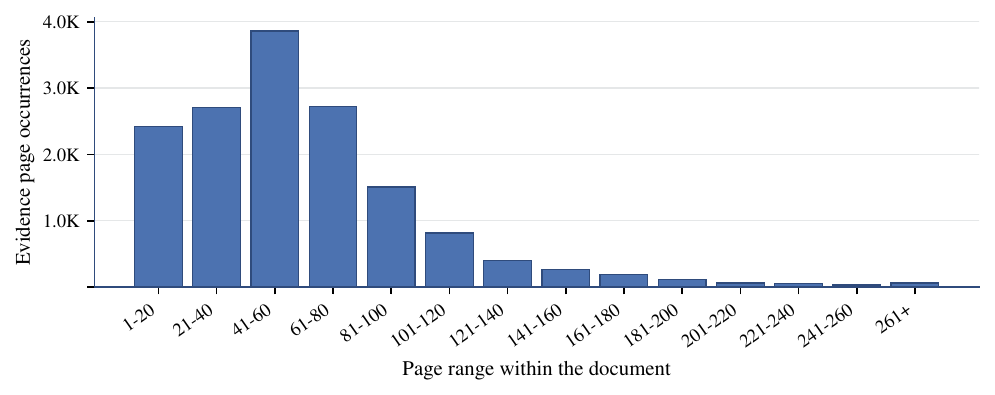}
  \caption{Distribution of evidence page occurrences across page ranges in annual reports.}
  \label{fig:qa-page-dist}
\end{figure}

\section{FinLongDocQA Dataset}
FinLongDocQA is a document-level financial QA dataset designed to evaluate both single-table and cross-table numerical reasoning in annual reports. It comprises 1{,}456 complete SEC filings from S\&P~500 companies for fiscal years 2022--2024, paired with 7{,}527 QA instances. Each instance is annotated with supporting evidence pages, enabling retrieval and verification in long-document settings.

We construct FinLongDocQA using a three-stage pipeline (Figure~\ref{fig:dataset_pipeline}). In \textbf{Stage~1 (QA generation)}, we parse each filing into a Markdown report, segment it into page-level chunks, and use an LLM to generate initial QA candidates from randomly sampled page ranges. In \textbf{Stage~2 (rule-based filtering)}, we automatically discard low-quality candidates by enforcing that each question has a valid numerical answer, executable Python code for the calculation, and evidence that requires multiple hops across pages rather than a single local lookup. In \textbf{Stage~3 (human review)}, we apply LLM-assisted checks and manual inspection to remove QA pairs that are off-topic, or poorly grounded, yielding the final dataset with traceable evidence pages suitable for cross-table, document-level financial reasoning.

\subsection{Data Collection}
We collect the corresponding PDF filings from the SEC EDGAR system\footnote{\texttt{https://www.sec.gov/search-filings}}. For each PDF, we apply MinerU~\cite{wang2024mineru} to extract tables, section titles, and text, converting each report into a Markdown file. This representation supports downstream indexing and retrieval while preserving the report content and much of its original layout. Document statistics are summarized in Table~\ref{tab:dataset_overview}.

\begin{figure*}[t]
    \centering
    \includegraphics[width=1\linewidth]{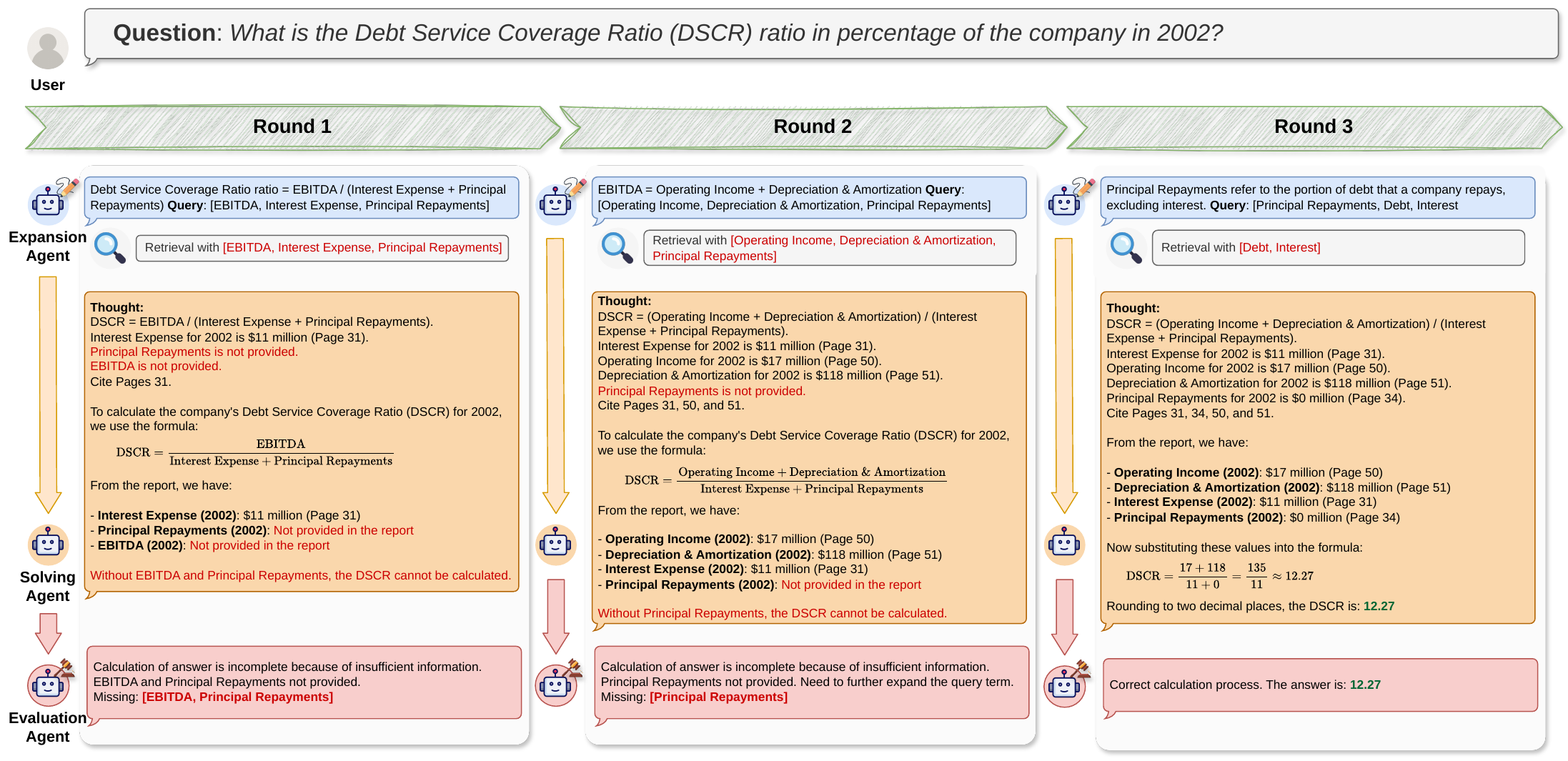}
    \caption{A running example of FinLongDocAgent showing multi-agent interaction across retrieval rounds.}
    \label{fig:method_finlongdocagent}
\end{figure*}

\subsection{QA Generation}

Manual expert annotation for document-level financial QA is costly, and recent work has increasingly turned to LLMs for automatic data construction. We follow this direction and use an LLM (\texttt{GPT-4o-mini}) to generate candidate QA pairs that require numerical computation across tables located on different pages of the report.

To create cross-page questions, we adopt a page-sampling strategy that selects disjoint subsets of pages from the early, middle, and late sections of each report. For each filing, we sample 40 pages from these regions and prompt the LLM to generate up to ten QA pairs using a template that includes chain-of-thought style reasoning and in-context examples. For each instance, the model produces not only the question and final numerical answer, but also (i) executable Python code that specifies how to compute the answer, (ii) a natural language reasoning trace, and (iii) the page indices that contain the supporting evidence. This initial QA generation procedure yields 13{,}375 raw QA pairs. 

Appendix Figure~\ref{fig:qa_generation_prompt} presents the prompt used for QA generation.


\subsection{Data Quality Assurance}
\label{sec:data_quality}
To ensure that FinLongDocQA require document-level financial reasoning, we apply a two-stage filtering process included of automatic rule-based checks followed by manual review. Table~\ref{tab:dataset_filtering} summarizes the filtering statistics.

\noindent\textbf{Rule-based filtering.}
We remove 4{,}120 questions whose answers can be derived from a single table, as these do not require cross-page or cross-table reasoning. We then execute the generated Python code associated with each QA pair and discard 657 instances with incorrect calculations. Finally, 268 QA pairs with incoherent, incomplete, or non-numeric answers are removed. After this stage, 8{,}330 QA pairs remain.



\noindent\textbf{Human review.}
The remaining QA pairs are further refined through a manual screening process conducted by two annotators with complementary backgrounds in computer science and finance. For \textit{inter-annotator agreement}, we adopt a conservative review protocol. Each QA pair is independently reviewed by both annotators. An instance is discarded if either annotator identifies issues such as ambiguity, overly explicit hints, inappropriate constraints, or lack of financial relevance. This protocol prioritizes precision and ensures that retained questions require well-defined reasoning.

We define the following criteria for discarding or modifying QA pairs:
\textbf{(i) Overly explicit hints:} questions that reveal formulas or definitions directly (e.g., explicitly writing ``ROA = Net income / Average assets'') are edited to avoid trivializing the reasoning process.
\textbf{(ii) Page-range constraints:} questions restricted to a narrow set of pages are removed if the answer would differ when considering the full report, as such constraints do not reflect realistic document-level analysis.
\textbf{(iii) Non-financial focus:} questions unrelated to financial reasoning (e.g., general workforce demographics) are discarded to maintain domain specificity.

This manual stage removes 803 QA pairs, yielding a final set of 7{,}527 finance-focused QA instances that require cross-table numerical reasoning grounded in complete annual reports. QA statistics are summarized in Table~\ref{tab:dataset_qa_stats}.

\subsection{Dataset Details}
To verify that FinLongDocQA reflects cross-table reasoning demands, we examine the distribution of pages from which evidence is drawn. Figure~\ref{fig:qa-page-dist} shows that evidence spans a wide range of locations within each report. Many QA pairs arise from the first 150 pages, where consolidated statements and management discussions typically appear, while a meaningful portion originates from later sections containing detailed segment analyses, regulatory disclosures, and risk factor notes. 

\section{FinLongDocAgent Method}

To establish a strong baseline for FinLongDocQA, we design an agentic RAG method tailored to its central challenge: retrieving and integrating dispersed financial evidence to perform cross-table numerical reasoning over long annual reports. Financial analysts typically work in an iterative manner to locate relevant line items and compute intermediate values. When discrepancies appear, they revisit earlier sections of the report to refine the analysis. Agent-based systems \cite{yao2022react, jin2025searchr, 10.1145/3768292.3770382} naturally reflect this workflow by coordinating stepwise reasoning with targeted search and verification.

Motivated by this observation, we introduce \textbf{FinLongDocAgent}, a multi-agent, multi-round RAG method built on AutoGen \cite{wu2024autogen}. The method retrieves evidence in multiple rounds, computes candidate answers from the retrieved content, and verifies whether all necessary components have been incorporated before returning a final result.

\subsection{Multi-Agent Architecture}
\label{sec:multi_agent_arch}
As shown in Figure~\ref{fig:method_finlongdocagent}, FinLongDocAgent consists of three agents that operate over multiple rounds. 

\noindent\textbf{Expansion Agent} converts the input question into a compact, formula-like expression that explicitly enumerates the financial line items needed for computation (e.g., mapping a ratio query to its numerator and denominator accounts). When feedback from previous rounds is available, the expansion input is augmented with the missing components identified by the evaluator, encouraging the agent to revise the expression to include unresolved accounts and their report-specific variants.

\noindent\textbf{Solving Agent} performs numerical reasoning using the retrieved context. It takes as input the original question, the expanded formula, and the retrieved report pages. To ensure traceability, it is instructed to justify each step with explicit \texttt{page\_number} references and the corresponding formula used. If the retrieved pages do not contain sufficient information to support a valid computation, the agent outputs a default value \{\{0\}\}.

\noindent\textbf{Evaluation Agent} verifies whether the proposed answer is complete and faithful to the question. It inspects the solving agent's output and checks two aspects: whether any critical components required for the computation are missing, and whether the accounts used match exactly those specified in the question. If the answer is complete, the evaluator returns a termination signal (i.e., \texttt{NONE}). Otherwise, it outputs a structured list of missing components together with common synonyms that may appear in annual reports. This output is fed back into the next expansion round, enabling targeted refinement of retrieval queries and reducing error propagation.

\subsection{Document Indexing}
Each annual report is provided in Markdown format and segmented into page-level chunks using explicit page delimiters. This page-level segmentation preserves the document’s native organization (e.g., statements, notes, and tables) while providing retrieval units that are sufficiently fine-grained for evidence localization. We study alternative chunking strategies in Appendix~\ref{app:chunk_ablation}. For each document, we construct a dense retrieval index once and reuse it across all questions associated with that report.

Overall, FinLongDocAgent combines multi-agent execution with iterative retrieval to better align with real financial analysis workflows. Appendix Figure~\ref{fig:finlongdocagent_prompts} presents the prompts used by FinLongDocAgent for each agent.

\section{Experimental Setup}
\label{sec:exp_setup}





\subsection{LLMs}
We evaluate both closed-source and open-source LLMs: \texttt{GPT-4o-mini}~\cite{achiam2023gpt} and \texttt{Gemini-3-Flash}\footnote{These models provide explicit reasoning capabilities and being enabled during inference}~\cite{google_gemini3} (closed-source), and \texttt{Qwen3-Thinking-30B}$^{2}$~\cite{yang2025qwen3} and \texttt{DeepSeek-v3.2}$^{2}$~\cite{liu2025deepseek} (open-source). All models are run in a zero-shot setting using a shared instruction template.


\subsection{Compared Methods}
We compare the following settings:

\noindent\textbf{LLM-only.}
We provide only the question to the model (without context), measuring how often models can answer from internal knowledge alone.

\noindent\textbf{Long-context prompting.}
We feed the model a report\footnote{Truncated to each model's maximum input length.} prefix without any retrieval component, to test its ability to reason over long inputs.

\noindent\textbf{Single-round RAG.}
We implement a standard retrieve-then-generate pipeline. The report is segmented into page-level chunks; we embed chunks and retrieve the top-$k$ ($k=30$) pages using a dense retriever (BGE~\cite{chen-etal-2024-m3}). 

\noindent\textbf{Graph-based RAG.}
We include GraphRAG~\cite{edge2024local} and HippoRAG2~\cite{gutierrez2025from}. For both methods, we construct report-level indices by following their original pipelines, and adopt \texttt{GPT-4o-mini} as the backbone LLM to ensure consistency with their procedures.

\begin{table}[t]
\centering
\small
\setlength{\tabcolsep}{4pt}
\renewcommand{\arraystretch}{1.05}
\begin{adjustbox}{max width=\columnwidth}
\begin{tabular}{lccc} 
\toprule
\textbf{Model} & \textbf{EM} & \textbf{Tol. Acc} & \textbf{F1} \\ 
\midrule
\textit{Without context} & \multicolumn{1}{l}{} & \multicolumn{1}{l}{} & \multicolumn{1}{l}{} \\
GPT-4o-mini & 0.12 & 0.13 & 0.56 \\
Gemini-3-Flash & 3.24 & 4.52 & 18.06 \\
Qwen3-Thinking 30B & 0.24 & 0.28 & 4.09 \\
DeepSeek-v3.2 & 1.18 & 1.47 & 10.39 \\ 
\midrule
\textit{With report (long-context prompting)} & \multicolumn{1}{l}{} & \multicolumn{1}{l}{} & \multicolumn{1}{l}{} \\
GPT-4o-mini & 7.88 & 11.73 & 15.06 \\
Gemini-3-Flash & 17.79 & 21.98 & 26.03 \\
Qwen3-Thinking 30B & 13.84 & 18.88 & 23.16 \\
DeepSeek-v3.2 & 13.45 & 18.51 & 22.25 \\ 
\midrule
\textit{With BGE retriever} & \multicolumn{1}{l}{} & \multicolumn{1}{l}{} & \multicolumn{1}{l}{} \\
GPT-4o-mini & 19.07 & 28.09 & 36.40 \\
Gemini-3-Flash & 33.03 & 41.93 & 49.90 \\
Qwen3-Thinking 30B & 27.21 & 36.47 & 42.43 \\
DeepSeek-v3.2 & 27.66 & 38.71 & 45.44 \\ 
\midrule
\textit{Graph-based RAG} & \multicolumn{1}{l}{} & \multicolumn{1}{l}{} & \multicolumn{1}{l}{} \\
GraphRAG (arXiv'24) & 12.14 & 18.11 & 25.41 \\
HippoRAG2 (ICML'25) & 15.12 & 22.77 & 30.83 \\ 
\midrule
\textit{Agentic RAG} & \multicolumn{1}{l}{} & \multicolumn{1}{l}{} & \multicolumn{1}{l}{} \\
WebDancer-30B (NeurIPS'25) & 22.60 & 31.80 & 38.27 \\
\textbf{FinLongDocAgent (Ours)} &  &  &  \\
\quad GPT-4o-mini & 20.97 & 30.51 & 36.97 \\
\quad Gemini-3-Flash & \textbf{41.34} & \textbf{43.54} & \textbf{51.29} \\
\quad Qwen3-Thinking 30B & 30.10 & 37.14 & 43.01 \\
\quad DeepSeek-v3.2 & 31.06 & 40.03 & 47.30 \\
\bottomrule
\end{tabular}
\end{adjustbox}
\caption{Main results on FinLongDocQA.}
\label{tab:model_comparison}
\vspace{-5pt}
\end{table}

\noindent\textbf{Agentic RAG.}
We include strong agentic retrieval \texttt{WebDancer-30B}~\cite{wu2025webdancer}. Since WebDancer is designed for web navigation, we adapt it by replacing web actions with retrieval over report chunks using the same BGE dense retriever.

\subsection{Retrieval Corpus and Chunking}
Each report is converted into Markdown and segmented into page-level chunks. For retrieval-based methods, chunks are embedded and indexed within the corresponding report.

\subsection{FinLongDocAgent Configuration}
FinLongDocAgent is implemented in AutoGen~\cite{wu2024autogen} and uses the same BGE retriever. We run at most 5 rounds, retrieving the top-$k$ chunks per round with $k{=}15$.

\subsection{Evaluation Metrics}
We report Exact Match (EM), Tolerance Accuracy (Tol.\ Acc), and token-level F1, following prior numerical QA benchmarks~\cite{zhu-etal-2021-tat}. The full evaluation protocol is provided in Appendix~\ref{app:eval_protocol}. 



\section{Experimental Results}
\label{sec:exp_results}

Table~\ref{tab:model_comparison} summarizes results on \textsc{FinLongDocQA}. Overall, access to document evidence is essential: all models perform near-zero without context, and accuracy increases substantially when evidence is provided via long-context prompting or retrieval.

\subsection{QA Results}

\noindent\textbf{No context.}
When only the question is provided, all models fail to reliably recover the gold numeric answers (Table~\ref{tab:model_comparison}). This confirms that \textsc{FinLongDocQA} cannot be solved by internal knowledge alone and requires grounding in report content.

\noindent\textbf{Long-context prompting.}
Prepending the report as a prefix improves all models, but performance remains limited (e.g., \texttt{Gemini-3-Flash}: 17.79 EM). This approach is susceptible to context rot, where salient evidence in later footnotes or dispersed tables is ignored as input length grows, highlighting the difficulty of reliable evidence localization in long financial reports.

\noindent\textbf{Single-round dense RAG.}
Single-round dense retrieval yields a large jump across backbones. With BGE retrieval, \texttt{Gemini-3-Flash} reaches 33.03 EM and \texttt{GPT-4o-mini} reaches 19.07 EM. These gains indicate that retrieving globally relevant pages mitigates missing-evidence failures that long-context prompting cannot address.

\noindent\textbf{Graph-based RAG.}
GraphRAG and HippoRAG2 underperform dense RAG methods. Annual reports contain many table-local quantities and implicit linkages that are not captured reliably by entity-relation graphs, so graph indices can lose necessary context around table cells.

\noindent\textbf{Agentic RAG.}
\texttt{WebDancer-30B} improves over long-context prompting but remains weaker than dense RAG methods, which may indicate limited generalization to long-document financial reasoning beyond its training domain. In contrast, FinLongDocAgent consistently outperforms single-round RAG across all evaluated backbones. For \texttt{Gemini-3-Flash}, FinLongDocAgent achieves the best overall performance, reaching 41.34 EM, 43.54 Tol.\ Acc., and 51.29 F1. These results suggest that iterative retrieval with explicit verification is particularly beneficial when required operands are scattered across multiple tables and notes.

\begin{table}[t]
\centering
\small
\setlength{\tabcolsep}{4pt}
\renewcommand{\arraystretch}{1.05}

\begin{adjustbox}{max width=\columnwidth}
\begin{tabular}{lccc}
\toprule
\textbf{Retriever} & \textbf{Recall@5} & \textbf{Recall@10} & \textbf{Recall@30} \\
\midrule
TF-IDF & 26.04 & 36.86 & 57.65 \\
BM25   & 22.58 & 31.91 & 51.12 \\
E5-Mistral-7B & 25.95 & 36.68 & 59.81 \\
BGE-M3    & \textbf{28.14} & \textbf{39.06} & \textbf{62.22} \\
\bottomrule
\end{tabular}
\end{adjustbox}
\caption{Retrieval effectiveness on FinLongDocQA.}
\label{tab:retriever_effect}
\end{table}

\begin{table}[t]
\centering
\begin{adjustbox}{max width=\columnwidth}
\begin{tabular}{lccc} 
\toprule
\multirow{2}{*}{\textbf{Model}} & \multicolumn{3}{c}{\textbf{Exact Match }} \\
 & \begin{tabular}[c]{@{}c@{}}\textit{Easy}\\(3,695)\end{tabular} & \begin{tabular}[c]{@{}c@{}}\textit{Medium}\\(3,721)\end{tabular} & \begin{tabular}[c]{@{}c@{}}\textit{Hard}\\(110)\end{tabular} \\ 
\hline
\textit{Without context} & \multicolumn{1}{l}{} & \multicolumn{1}{l}{} & \multicolumn{1}{l}{} \\
GPT-4o-mini & 0.19 & 0.00 & 0.00 \\
Gemini-3-Flash & 3.75 & 2.50 & 0.00 \\
Qwen3-Thinking 30B & 0.36 & 0.04 & 0.00 \\
DeepSeek-v3.2 & 1.24 & 1.10 & 0.91 \\ 
\midrule
\textit{With report (long-context prompting)} & \multicolumn{1}{l}{} & \multicolumn{1}{l}{} & \multicolumn{1}{l}{} \\
GPT-4o-mini & 8.92 & 6.39 & 0.00 \\
Gemini-3-Flash & 19.40 & 15.47 & 6.36 \\
Qwen3-Thinking 30B & 15.87 & 10.69 & 5.45 \\
DeepSeek-v3.2 & 14.21 & 12.31 & 9.09 \\ 
\midrule
\textit{With BGE retriever} & \multicolumn{1}{l}{} & \multicolumn{1}{l}{} & \multicolumn{1}{l}{} \\
GPT-4o-mini & 19.45 & 18.63 & 13.64 \\
Gemini-3-Flash & 34.33 & 31.14 & 24.55 \\
Qwen3-Thinking 30B & 29.50 & 23.73 & 15.55 \\
DeepSeek-v3.2 & 29.19 & 25.60 & 13.64 \\ 
\midrule
\textit{Graph-based RAG} & \multicolumn{1}{l}{} & \multicolumn{1}{l}{} & \multicolumn{1}{l}{} \\
GraphRAG (arXiv'24) & 13.27 & 10.40 & 7.27 \\
HippoRAG2 (ICML'25) & 16.42 & 13.12 & 9.09 \\ 
\midrule
\textit{Agentic RAG} & \multicolumn{1}{l}{} & \multicolumn{1}{l}{} & \multicolumn{1}{l}{} \\
WebDancer-30B (NeurIPS'25) & 23.37 & 21.71 & 11.82 \\
\textbf{FinLongDocAgent (Ours)} &  &  &  \\
\quad GPT-4o-mini & 22.68 & 18.28 & 14.55 \\
\quad Gemini-3-Flash & \textbf{43.99} & \textbf{36.99} & \textbf{35.45} \\
\quad Qwen3-Thinking 30B & 33.93 & 23.89 & 20.00 \\
\quad DeepSeek-v3.2 & 34.01 & 26.32 & 22.46 \\
\bottomrule
\end{tabular}
\end{adjustbox}
\caption{Exact Match (EM) performance on FinLongDocQA across different difficulty levels, categorized by the number of table evidence sources: Easy ($\leq$1 table), Medium (2 tables), and Hard ($\geq$3 tables).}
\label{tab:model_comparison_with_difficulty}
\vspace{10pt}
\end{table}

\begin{table}[t]
\centering
\small
\setlength{\tabcolsep}{6pt}
\renewcommand{\arraystretch}{1.05}
\begin{adjustbox}{max width=\columnwidth}
\begin{tabular}{lccc}
\toprule
\textbf{Setting} & \textbf{EM} & \textbf{Tol.\ Acc} & \textbf{F1} \\
\midrule
Solving Agent & 33.03 & 41.93 & 49.90 \\
\quad+ Expansion Agent & 36.74 & 42.12 & 50.01 \\
\quad\quad+ Evaluation Agent & \textbf{41.34} & \textbf{43.54} & \textbf{51.29} \\
\bottomrule
\end{tabular}
\end{adjustbox}
\caption{Ablation study of FinLongDocAgent components (LLM: Gemini-3-Flash).}
\label{tab:ablation_finlongdocagent}
\end{table}

\subsection{Retriever Effectiveness.}
Table~\ref{tab:retriever_effect} reports Recall@$k$ against the gold evidence pages. BGE achieves the highest recall at all cutoffs. While dense retrievers generally outperform sparse baselines, Recall@30 remains far from saturated, indicating that evidence localization is still a major bottleneck in FinLongDocQA.

\begin{table}[t]
\centering
\small
\setlength{\tabcolsep}{0pt}
\renewcommand{\arraystretch}{1.12}

\definecolor{mutedblue}{RGB}{45,90,160}
\definecolor{mutedred}{RGB}{170,70,70}

\begin{tabular}{p{0.98\columnwidth}}
\toprule
\textbf{Error Analysis} \\
\midrule

\rowcolor[rgb]{0.93,0.93,0.93}
\textbf{\textit{Retrieval Error (62\%)}} \\
\parbox[t]{0.98\columnwidth}{%
\textbf{Q:} What was the 2023 net sales per facility in Europe, the Middle East \& Africa?\par
\textcolor{mutedblue}{\textbf{G:}} 518.75\par
\textbf{Retrieved:} 29, 32, 36, 37, 39 \textit{(+10 more)}\par
\textbf{Failure cue:} \textcolor{mutedred}{Missing required operands} (EMEA net sales; \# facilities).\par
\textcolor{mutedred}{\textbf{P:}} \textit{Retrieved pages do not contain} (2023 EMEA Net Sales) / (2023 EMEA \# Facilities).%
}\\
\midrule

\rowcolor[rgb]{0.93,0.93,0.93}
\textbf{\textit{Evidence Utilization Error (19\%)}} \\
\parbox[t]{0.98\columnwidth}{%
\textbf{Q:} What was the total lease liability per proved BoE of reserves at year-end 2022?\par
\textcolor{mutedblue}{\textbf{G:}} 0.06 \quad \textbf{(GT: p.101, p.105)}\par
\textbf{Retrieved:} 101, 105, 104, 108, 107 \textit{(+10 more)}\par
\textbf{Failure cue:} Gold pages are retrieved, but the model \textcolor{mutedred}{uses reserves from the wrong page}.\par
\textcolor{mutedred}{\textbf{P:}}%
$\frac{296+33+584+182\;(\textcolor[rgb]{0,0.5,0}{p.101})}{4{,}238\;(\textcolor{mutedred}{p.107})}=0.26$%
}\\
\midrule

\rowcolor[rgb]{0.93,0.93,0.93}
\textbf{\textit{Value Extraction Error (11\%)}} \\
\parbox[t]{0.98\columnwidth}{%
\textbf{Q:} What is the absolute difference between 2022 underlying net income available to common stockholders and GAAP net income?\par
\textcolor{mutedblue}{\textbf{G:}} 352 \quad \textbf{(GT: p.16, p.17)}\par
\textbf{Retrieved:} 16, 17, 15, 14, 18 \textit{(+10 more)}\par
\textbf{Failure cue:} Evidence is retrieved and used, but the model \textcolor{mutedred}{extracts a wrong GAAP value}.\par
\textcolor{mutedred}{\textbf{P:}} $|2312-\textcolor{mutedred}{2073}|=\textcolor{mutedred}{239}$ \quad (should be 352)%
}\\
\midrule

\rowcolor[rgb]{0.93,0.93,0.93}
\textbf{\textit{Reasoning / Calculation Error (8\%)}} \\
\parbox[t]{0.98\columnwidth}{%
\textbf{Q:} What was the cash conversion ratio in fiscal 2023?\par
\textcolor{mutedblue}{\textbf{G:}} 0.82 \quad \textbf{(GT: p.71, p.74)}\par
\textbf{Retrieved:} 71, 74, 75, 64, 60 \textit{(+10 more)}\par
\textbf{Failure cue:} Operands appear correct, but the model \textcolor{mutedred}{computes the ratio incorrectly}.\par
\textcolor{mutedred}{\textbf{P:}}%
$\text{FCF}=640-183=457,$\allowbreak\ 
$\text{CCR}=\frac{\textcolor{mutedred}{457}}{783}=\textcolor{mutedred}{0.584}$%
}\\
\bottomrule
\end{tabular}

\caption{Error analysis of FinLongDocAgent. \textbf{Q}: question, \textcolor{mutedblue}{\textbf{G}}: golden answer, \textcolor{mutedred}{\textbf{P}}: model prediction.}
\label{tab:error_analysis}
\vspace{5pt}
\end{table}

\subsection{Detailed Analysis}
\label{subsec:ablation}
\noindent\textbf{Ablation Study.}
Table~\ref{tab:ablation_finlongdocagent} ablates FinLongDocAgent components using \texttt{Gemini-3-Flash}. Adding the query expansion agent improves EM from 33.03 to 36.74, suggesting that metric-aware reformulation increases evidence recall. Adding the evaluation agent yields the largest gain, raising EM to 41.34 and improving Tol.\ Acc to 43.54. Qualitatively, the evaluation agent reduces failures due to missing operands, unit/scale mismatches, and inconsistent intermediate computations by triggering additional retrieval rounds when needed.



\noindent\textbf{Question Difficulty Evaluation.}
We categorize questions based on the number of unique evidence tables required to derive the answer: Easy ($\leq$1), Medium (2), and Hard ($\geq$3), containing 3,695, 3,721, and 110 questions, respectively. Table~\ref{tab:model_comparison_with_difficulty} reports Exact Match (EM) performance across these difficulty levels. Performance consistently decreases as the number of required tables increases, highlighting the difficulty of cross-table numerical reasoning in long financial documents.

\noindent\textbf{Error Analysis.}
Table~\ref{tab:error_analysis} categorizes FinLongDocAgent failures into four types. The dominant source of error is \textit{retrieval failure} (62\%), where at least one required page is not retrieved. The remaining errors arise even when relevant pages are present in the retrieved set: \textit{evidence utilization errors} (19\%) occur when the model utilizes an incorrect page among the retrieved candidates, \textit{value extraction errors} (11\%) result from using wrong table entries, and \textit{reasoning/calculation errors} (8\%) are due to incorrect arithmetic or an incorrect formulation. Overall, these findings suggest that improving evidence recall is necessary but not sufficient; robust evidence utilization and accurate extraction from retrieved tables remain critical.

\subsection{Token-Budget Fairness}
\label{subsec:token_budget}

A potential concern is that the improvements of FinLongDocAgent may arise from accessing a larger total context budget rather than from its agentic design. In our implementation, FinLongDocAgent retrieves up to 15 chunks per round for a maximum of 5 rounds, yielding at most 75 retrieved chunks.

To control for this factor, we evaluate a single-round RAG with the same maximum retrieval budget (75 chunks). With \texttt{GPT-4o-mini}, the standard single-round RAG baseline using 30 retrieved chunks achieves 19.07 EM. Increasing the retrieval budget to 75 chunks reduces performance to 16.64 EM, likely due to increased noise in the context.

In contrast, FinLongDocAgent achieves 20.97 EM on \texttt{GPT-4o-mini} and 41.34 EM on \texttt{Gemini-3-Flash}. These results suggest that simply increasing the retrieval budget does not improve performance. Instead, the gains of FinLongDocAgent stem from iterative query rewriting and verification, which progressively filter irrelevant context and ensure that required operands are collected before numerical reasoning.

\section{Conclusion}
We introduced \textbf{FinLongDocQA}, a document-level QA dataset for both single-table and \emph{cross-table} numerical reasoning in annual reports. FinLongDocQA is challenging because models must locate relevant operands across multiple pages and heterogeneous formats (e.g., tables and textual descriptions), and then perform multi-step computations.
Experiments on both closed-source and open-source LLMs show that performance remains limited. To address this challenge, we propose \textbf{FinLongDocAgent}, a multi-agent, multi-round RAG framework with iterative retrieval and verification, which consistently outperforms single-round and other baseline methods. Our error analysis further reveals that retrieval remains the primary bottleneck in document-grounded financial numerical QA.
Looking forward, we hope FinLongDocQA will foster research on more robust long-document evidence discovery, structure-aware grounding over tables, and reliable executable numerical reasoning.

\newpage
\section*{Limitations}
Our benchmark and baselines involve several practical design choices. Converting filings into Markdown enables scalable indexing but may not preserve all layout cues, such as multi-column reading order, which can affect value extraction in some cases. FinLongDocAgent improves robustness through iterative retrieval and explicit verification, at the cost of increased inference overhead and continued dependence on retrieval quality. Finally, FinLongDocQA is constructed from S\&P~500 annual reports from 2022--2024; extending evaluation to other time periods, modalities, and markets remains an important direction for future work.


\section*{Ethical Considerations}
FinLongDocQA is constructed from publicly available SEC filings and does not involve private or user-generated data. Although filings may contain names of company executives as part of the public record, we do not curate questions that target individuals. 
Models evaluated on this benchmark may produce incorrect retrievals or numerical reasoning; if deployed without appropriate safeguards, such errors could lead to misleading financial interpretations. Accordingly, FinLongDocQA is intended solely for research evaluation, and its scope is limited to S\&P~500 filings from 2022--2024.
We use an LLM (\texttt{GPT-4o-mini}) to generate initial QA candidates. To mitigate risks from synthetic artifacts, we apply rule-based filtering, verify calculations, and perform careful manual review.

\bibliography{custom}

\clearpage
\appendix
\section{Appendix}
\label{sec:appendix}

\noindent\textbf{Outline.}
Section~\ref{app:data_details} details dataset construction.
Section~\ref{app:method_details} provides FinLongDocAgent implementation details.
Section~\ref{app:eval_protocol} defines evaluation metrics.
Section~\ref{app:extra_repro} reports additional experiments, and case studies.

\subsection{Dataset Construction Details}
\label{app:data_details}

\noindent\textbf{Document preprocessing and table extraction.}
We convert each filing PDF into a page-level Markdown document using MinerU. Each page chunk contains:
(i) running text extracted from the PDF,
(ii) detected tables serialized into Markdown, and
(iii) the page identifier.
We keep page boundaries to enable evidence supervision and recall evaluation.

\noindent\textbf{QA generation prompt and post-processing.}
We generate candidate QA pairs using \texttt{GPT-4o-mini}. The prompt instructs the model to propose a finance-relevant metric whose computation requires evidence from multiple pages, output the final numerical answer, produce executable Python code that computes the answer, and specify the page indices containing the supporting evidence. Figure~\ref{fig:qa_generation_prompt} presents the prompt we used.

\noindent\textbf{Rule-based filtering heuristics.}
This paragraph details the automatic rules used in Stage~2 (Section~\ref{sec:data_quality}).
\textbf{Multi-page / multi-hop constraint:} We require at least two distinct evidence pages. We additionally remove cases where all operands appear in a single table on one page by checking whether the referenced values are co-located in the same extracted table block.
\textbf{Answer type constraint:} We discard questions with non-numeric answers or answers dependent on unverifiable textual judgments (e.g., ``Is the company optimistic?'').
\textbf{Executable program constraint:} We execute the provided program and remove instances where:
(i) the program fails, (ii) the result is not numeric, or (iii) the program output deviates from the stated answer beyond a tolerance threshold.
\textbf{Incoherence constraint:} We remove instances where the question is ambiguous (missing year/segment) or the reasoning trace contradicts the program.

\noindent\textbf{Human review protocol.}
Two annotators conduct manual review on a FinLongDocQA.
The review follows a checklist:
(1) financial relevance,
(2) clear definition of the requested quantity (year, segment, unit),
(3) evidence sufficiency (pages truly support operands),
(4) absence of overly explicit hints (e.g., directly stating full formula),
(5) no dependence on outside knowledge.
Disagreements are resolved by discussion.

\begin{figure}[t]
    \centering
    \includegraphics[width=0.45\textwidth]{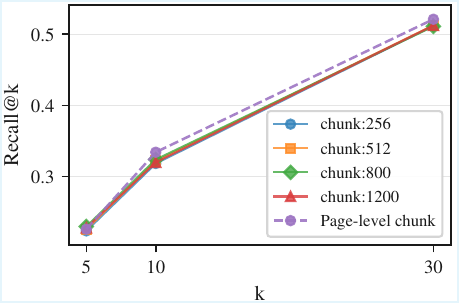}
    \vspace{-8pt}
    \caption{Recall@k for different chunk sizes.}
    \label{fig:chunk_k_ablation}
\end{figure}

\begin{table}[t]
\centering
\small
\begin{tabular}{lcc} 
\toprule
\textbf{Methods} & \begin{tabular}[c]{@{}c@{}}\textbf{Prompting}\\\textbf{Design}\end{tabular} & \textbf{FinQA$^{\dagger}$} \\ 
\midrule
FinQAPT~\citep{10.1145/3677052.3698682} & Few-shot & 62.0 \\
FinLongDocAgent (ours) & Zero-shot & \textbf{66.2} \\
\bottomrule
\end{tabular}
\vspace{2pt}
\caption{Comparison with a prior SOTA method on FinQA. $^{\dagger}$Modified setting where the full document is provided as input.}
\label{tab:finqapt_finqa}
\end{table}


\subsection{FinLongDocAgent: Additional Method Details}
\label{app:method_details}

\noindent\textbf{Implementation details for retrieval.}
\textbf{Indexing:} We build a per-report index over page chunks. For dense retrieval, we embed each chunk using BGE-M3 and retrieve via inner-product similarity.
\textbf{Query formulation:} For single-round RAG, the query is the question text. For FinLongDocAgent, the query expansion agent produces metric-aware queries (Section~\ref{sec:multi_agent_arch}), e.g., adding synonyms and likely line-item names found in reports (``net sales'' $\rightarrow$ ``revenue'', ``interest expense'' $\rightarrow$ ``interest and other expense'').
\textbf{Top-$k$ choices:} Single-round RAG uses $k{=}30$. FinLongDocAgent uses $k{=}15$ per round for up to five rounds. Figure~\ref{fig:finlongdocagent_prompts} presents the prompts used for each agent.


\subsection{Evaluation Metrics}
\label{app:eval_protocol}

We report EM, Tolerance Accuracy (Tol.\ Acc), and token-level F1.
\textbf{Normalization:} Predictions and gold answers are normalized by removing commas, currency symbols, and whitespace, and converting parentheses negatives (e.g., ``(1,234)'') to $-1234$.
Percent answers are converted to either fraction or percentage consistently based on question cues (we store an explicit flag when available).

\noindent\textbf{Exact Match (EM):} EM is 1 if the normalized prediction string matches the normalized gold answer string exactly, otherwise 0.

\noindent\textbf{Tolerance Accuracy (Tol.\ Acc):} A prediction $\hat{a}$ is correct if: $|\hat{a}-a| \le a_{\text{tol}} + r_{\text{tol}} \cdot \max(|a|,\epsilon),$ where $a_{\text{tol}}{=}10^{-4}$, $r_{\text{tol}}{=}10^{-3}$, and $\epsilon{=}10^{-12}$. 

\noindent\textbf{Token-level F1:} We tokenize normalized predictions and references using whitespace and punctuation delimiters to compute token-level F1.

\subsection{Additional Experiments and Reproducibility}
\label{app:extra_repro}

\noindent\textbf{Effect of Chunk Granularity.}
\label{app:chunk_ablation}
We ablate chunk granularity for BM25 retrieval while keeping evaluation page-grounded.
We vary chunk length (\{256, 512, 800, 1200\} tokens), chunking \emph{within each page}.
We report page-level Recall@\{5,10,30\} and include a page-level chunk method that indexes each page as one unit.
As shown in Figure~\ref{fig:chunk_k_ablation}, page-level chunking slightly exceeds sub-page chunking across $k$, suggesting that treating a page as the retrieval unit may better preserve page-level evidence structure (e.g., keeping tables intact rather than fragmenting rows/cells across chunks).

\noindent\textbf{Comparison with FinQAPT.}
FinQAPT~\citep{10.1145/3677052.3698682} reports strong performance on FinQA~\cite{chen-etal-2021-finqa}, but the system is not publicly released, which limits its evaluation on FinLongDocQA. As a reference point, we evaluate FinLongDocAgent on FinQA under the \emph{full-document} input setting (i.e., using complete reports rather than pre-selected evidence), following the protocol described in FinQAPT. As shown in Table~\ref{tab:finqapt_finqa}, FinLongDocAgent achieves 66.2 EM on FinQA, outperforming FinQAPT's reported 62.0 EM. Our approach operates in a zero-shot setting without curated exemplars, whereas FinQAPT relies on dynamic few-shot prompting.


\noindent\textbf{Full prompt templates.}
Figures~\ref{fig:qa_generation_prompt}, \ref{fig:finlongdocagent_expansion_prompt}, \ref{fig:finlongdocagent_solving_prompt}, and \ref{fig:finlongdocagent_eval_prompt} show the full prompt templates used for (i) QA generation and (ii) each agent role in FinLongDocAgent, respectively.

\newpage
\begin{figure*}[t]
\centering
\begin{tcolorbox}[qapromptbox]
\begin{Verbatim}[fontsize=\footnotesize, breaklines=true, breakanywhere=true]
You are a seasoned financial analyst. Your task is to read a company's annual report (provided below in Markdown) and generate challenging, multi-page numerical reasoning questions and answers.
- For each QA pair, first **think step by step** about which line items you need and on which pages they appear.
- Show your chain-of-thought (labeled “Thought: …”) to justify each calculation.
- Then give the final answer in a clear formulaic layout.

### Examples
Q1: What is the Inventory turnover ratio in percentage of the company in 2002?
A1: Inventory turnover ratio = COGS cost of goods sold / average inventory = 4139/[(45+11)/2] = 147.82

Q2: What is the Debt Service Coverage Ratio (DSCR) ratio in percentage of the company in 2002?
A2: Debt Service Coverage Ratio (DSCR) ratio = EBITDA earnings before interest,taxes,depreciation and amortization/ (Interest Expense+Principal Repayments) = 17 / (11+0) = 1.55

Q3: What is the Altman Z-Score in percentage of the company in 2002?
A3: Altman Z-Score = 1.2*(Working Capital/Total Assets) + 1.4*(Retained Earnings/Total Assets) + 3.3*(EBITDA/Total Assets) + 0.6*(Market Value of Equity/Total Liabilities) + 1.0*(Sales/Total Assets) = 1.2x(3730/6298) + 1.4x(2325/6298) + 3.3x(17/6298) + 0.6x(4925/2203) + 1.0x(5742/6298) = 3.49

### Now: Generate **10** new QA pairs in this exact format, each requiring data from at least two different pages of the provided report.
Be sure to:
- Prepend each reasoning with “Thought:”
- Cite the page numbers in your chain-of-thought.
- Show each formula calculation step by step.
- Deliver the final answer as python code based on the formula under the concept of rounding to two decimal places.
- Run the code to get the final answer.
- Data must strictly come from *at least 2 to 3 different pages*.
- If the page is already used in the previous question, it is not allowed to be used again.
- Present your response in a strictly structured format through the json format below:
{
  "id": "unique_id_for_this_qa_pair",
  "company": "Company Name",
  "year": "Year of the report",
  "question": "What is the ...?",
  "type": "If the data sources are from table, please answer 'table'; if the data sources are from text, please answer 'text'; if the data sources are from both table and text, please answer 'mixed'",
  "thoughts": "Thought: ...",
  "page_numbers": [1, 2, 3],
  "python_code": "Based on the formula under the concept of strictly rounding to two decimal places.",
  "answer": "Numerical answer here"
}
IMPORTANT: Return only the raw JSON array—no Markdown fences!
\end{Verbatim}
\end{tcolorbox}
\caption{Full prompt template used for QA generation.}
\label{fig:qa_generation_prompt}
\vspace{20pt}
\end{figure*}

\begin{figure*}[t]
\centering

\begin{subfigure}{0.98\textwidth}
\centering
\begin{tcolorbox}[qapromptbox,title={FinLongDocAgent Expansion Agent System Prompt}]
\begin{Verbatim}[fontsize=\footnotesize, breaklines=true, breakanywhere=true]
You are an expert at financial text understanding.
Given a user's question about a financial ratio in an annual report, think of which accounts will feed into that formula, and *only* return the complete formula as a whole—no explanations.
\end{Verbatim}
\end{tcolorbox}
\caption{Expansion Agent.}
\label{fig:finlongdocagent_expansion_prompt}
\end{subfigure}

\vspace{4pt}

\begin{subfigure}{0.98\textwidth}
\centering
\begin{tcolorbox}[qapromptbox,title={FinLongDocAgent Solving Agent System Prompt}]
\begin{Verbatim}[fontsize=\footnotesize, breaklines=true, breakanywhere=true]
You are a helpful assistant that answers financial QA based on the provided annual report in Markdown form.
You should clearly show your data source through page_number and formula to justify each calculation.
Your final numerical answer must be rounded to two decimal places and enclosed in double curly braces {{}} with no extra text or units.
If the information is insufficient, put 0 inside the {{}}.

Given a user's question about a financial ratio in an annual report, think of which line items (accounts) will feed into that formula, and *only* return the complete formula as a whole-no explanations.
\end{Verbatim}
\end{tcolorbox}
\caption{Solving Agent.}
\label{fig:finlongdocagent_solving_prompt}
\end{subfigure}

\vspace{4pt}

\begin{subfigure}{0.98\textwidth}
\centering
\begin{tcolorbox}[qapromptbox,title={FinLongDocAgent Evaluation Agent System Prompt}]
\begin{Verbatim}[fontsize=\footnotesize, breaklines=true, breakanywhere=true]
You are an expert assistant that inspects a generated answer against the question.
You have two tasks:
1) If the answer is missing any critical components needed to compute the numeric result, return a comma-separated list of exactly those missing components (e.g., 'net income', 'total assets'). For each missing component, also include common synonyms or variant phrasings that might appear in the report (e.g., for 'COGS': 'cost of goods sold', 'cost of sales').
2) If the answer is not using 'exactly' the same accounts as the question stated (e.g., 'Consolidated other assets' vs 'Condensed other assets'), return a comma-separated list of the exact account names mentioned in the question. For each incorrect account, also include common synonyms or variant phrasings that might appear in the report (e.g., for 'COGS': 'cost of goods sold', 'cost of sales').
If nothing is missing and the answer uses exactly the same accounts as in the question, reply with 'NONE'.
\end{Verbatim}
\end{tcolorbox}
\caption{Evaluation Agent.}
\label{fig:finlongdocagent_eval_prompt}
\end{subfigure}

\caption{System prompts used in FinLongDocAgent.}
\label{fig:finlongdocagent_prompts}
\end{figure*}

\end{document}